\pdfoutput=1

\documentclass[11pt]{article}

\usepackage[final]{acl}

\usepackage{times}
\usepackage{latexsym}

\usepackage[T1]{fontenc}

\usepackage[utf8]{inputenc}
\usepackage[shortlabels]{enumitem}

\usepackage{microtype}
\usepackage{graphicx}
\usepackage{linguex}
\usepackage{subcaption}
\usepackage{amssymb}

\usepackage{multicol}
\usepackage{multirow} 
\usepackage{booktabs}

\RequirePackage{etoolbox}
\RequirePackage{xcolor}
\definecolor{Gray}{gray}{0.9}
\definecolor{cb-black}      {RGB}{ 0,   0,   0}
\definecolor{cb-blue-green} {RGB}{ 0,  073,  073}
\definecolor{cb-green-sea}  {RGB}{ 0, 146, 146}
\definecolor{cb-rose}       {RGB}{255, 109, 182}
\definecolor{cb-salmon-pink}{RGB}{255, 182, 119}
\definecolor{cb-purple}     {RGB}{ 73,   0, 146}
\definecolor{cb-blue}       {RGB}{ 0, 109, 219}
\definecolor{cb-lilac}      {RGB}{182, 109, 255}
\definecolor{cb-blue-sky}   {RGB}{109, 182, 255}
\definecolor{cb-blue-light} {RGB}{182, 219, 255}
\definecolor{cb-burgundy}   {RGB}{146,   0,   0}
\definecolor{cb-brown}      {RGB}{146,  73,   0}
\definecolor{cb-clay}       {RGB}{219, 209,   0}
\definecolor{cb-green-lime} {RGB}{ 36, 255,  36}
\definecolor{cb-yellow}     {RGB}{255, 255, 109}

\usepackage{stfloats}

\usepackage{inconsolata}
\usepackage{hyperref}
\usepackage{fontawesome5}
\newcommand{\aipom}[0]{\raisebox{-.25\height}{\includegraphics[width=.06\textwidth]{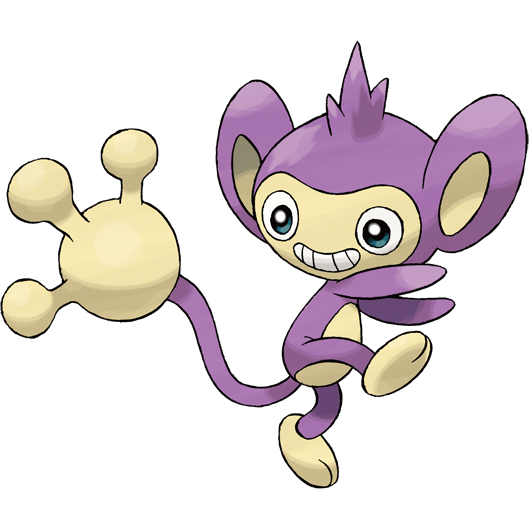}}} 

\newcommand{\man}[0]{\includegraphics[width=.02\textwidth]{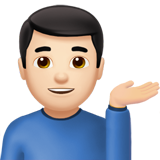}}

\newcommand{\lifting}[0]{\includegraphics[width=.02\textwidth]{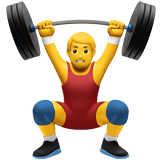}}

\newcommand{\ball}[0]{\includegraphics[width=.02\textwidth]{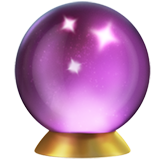}}

\newcommand{\ballot}[0]{\includegraphics[width=.02\textwidth]{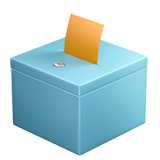}}

\newcommand{\machine}[0]{\includegraphics[width=.02\textwidth]{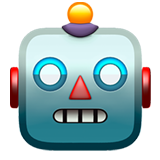}}

\title{\aipom AIpom at SemEval-2024 Task 8: Detecting AI-produced Outputs in M4}

\author{ Alexander Shirnin \textsuperscript{\faCrow}, 
        Nikita Andreev\textsuperscript{\faTree} \\
     \textbf{Vladislav Mikhailov}\textsuperscript{\faSnowflake},
    \textbf{Ekaterina Artemova}\textsuperscript{\faPaw} \\
    \textsuperscript{\faCrow}HSE University,
    \textsuperscript{\faTree}CAIT and Applied AI Institute \\
    \textsuperscript{\faSnowflake}University of Oslo,
    \textsuperscript{\faPaw}Toloka AI  \\
    \small{
    \textbf{Correspondence:} \href{mailto:ashirnin@hse.ru}{\texttt{ashirnin@hse.ru}}
}
}

\begin{document}
\maketitle
\begin{abstract}
This paper describes AIpom, a system designed to detect a boundary between human-written and machine-generated text (SemEval-2024 Task 8, Subtask C: Human-Machine Mixed Text Detection). We propose a two-stage pipeline combining predictions from an instruction-tuned decoder-only model and encoder-only sequence taggers. AIpom is ranked second on the leaderboard while achieving a Mean Absolute Error of 15.94.  Ablation studies confirm the benefits of pipelining encoder and decoder models, particularly in terms of improved performance.
\end{abstract}

\section{Introduction}
SemEval-2024 Task 8~\cite{task8semeval} focuses on multigenerator, multidomain, and multilingual machine-generated text detection based on the M4 corpus~\cite{wang-etal-2024-m4}. The shared task offers three subtasks, which correspond to standard task formulations in the rapidly developing field of artificial text detection~\cite{jawahar-etal-2020-automatic,uchendu2023reverse}: (A)~classifying if a given text in a particular language is human-written or machine-generated, (B)~attributing the author of a given text, and (C)~detecting a boundary between human-written and machine-generated text. Developing generalizable solutions to these problems helps mitigate the risks of misusing generative language models (LMs) for malicious purposes~\cite{weidinger2022taxonomy} and improve human performance in identifying AI-produced content~\cite{gehrmann-etal-2019-gltr}.

This paper proposes AIpom\footnote{AIpom is named after a simian pokémon \textit{aipom}, and stands for detecting \textbf{AI}-\textbf{p}roduced \textbf{o}utputs in \textbf{M}4.}, a novel method for human-machine mixed text detection (Subtask C). The boundary detection setup aligns with common user scenarios for applying generative LMs in practice, e.g., text continuation, creative writing, and story generation. The standard approach to this task is training a linear classifier or a regression model over encoder representations~\cite{cutler2021automatic,dugan2023real}. In contrast, AIpom leverages a pipeline of decoder and encoder models to detect machine-generated text, utilizing them sequentially. AIpom takes second out of 33 participating teams on the Subtask C leaderboard by achieving a Mean Absolute Error (MAE) of 15.94 on the official evaluation set. After the official evaluation phase, we develop a better-performing solution with an MAE score of 15.21.

Our ablation studies confirm that using decoder or encoder models individually leads to lower performance. Thus, employing the pipeline of decoder and encoder models proves to be an effective solution. Additionally, these studies highlight domain shift issues, as there is a significant score disparity between the development and official evaluation sets. Future efforts should focus on enhancing the AIpom robustness with respect to the text domain and text generator. The codebase and models are publicly released\footnote{\href{https://github.com/25icecreamflavors/AIpom}{\texttt{github.com/25icecreamflavors/AIpom}}}.

\section{Background}
The M4 corpus consists of human-written and machine-generated texts in six languages (English, Chinese, Russian, Urdu, Indonesian, and Arabic) across various domains, ranging from Wikipedia to academic peer reviews. The generative LMs include the OpenAI models (ChatGPT and text-davinci-003), LLaMA-1~\cite{touvron2023llama}, FLAN-T5~\cite{chung2022scaling}, Cohere, Dolly-v2\footnote{\href{https://huggingface.co/databricks/dolly-v2-12b}{\texttt{hf.co/databricks/dolly-v2-12b}}}, and BLOOMZ~\cite{muennighoff2022crosslingual}. The organizers provide 3649, 505, and 11123 dataset instances in Subtask C training, development, and official evaluation sets, respectively. 

\paragraph{Task Formulation} Human-machine mixed text detection requires predicting the index corresponding to the first machine-generated word, as shown below:

\begin{itemize}
    \item \texttt{text:} ``\man\textcolor{cb-black}{We have} added a 2+ page \machine\textcolor{cb-burgundy}{discussion on the experimental results, highlighting the superiority of the ARC-based models and their impact on the field of deep learning.}''
    \item \texttt{label: 6}
\end{itemize}

\paragraph{Performance Metric} MAE measures the absolute distance between the predicted word and the word where the human-machine transition occurs.

\section{AIpom}\label{sec:aipom}

First, we overview the AIpom pipeline. Next, we detail the fine-tuning procedures for encoder and decoder models.

\begin{figure*}[htp!]
    \centering
    \includegraphics[width = 0.9\textwidth]{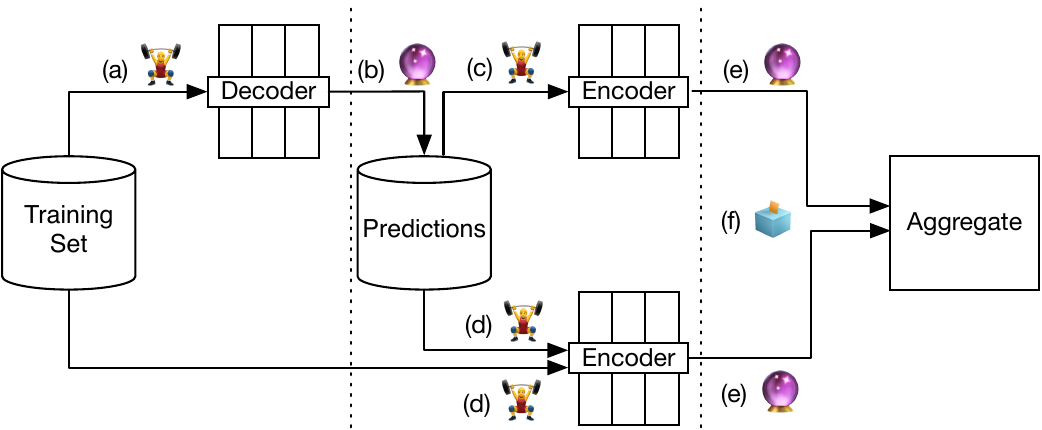}
    \caption{The AIpom pipeline involves fine-tuning decoder and encoder models to predict change points between the human-written and machine-generated text. This process includes fine-tuning the decoder, predicting change points, fine-tuning two encoders, and aggregating predicted change points. \lifting \ stands for fine-tuning a language model, \ball~--~predicting with the language model, \ballot \ -- for aggregating the predictions by averaging.}
    \label{fig:pipeline}
\end{figure*}

\paragraph{Overview} The AIpom pipeline (see \autoref{fig:pipeline}) consists of multiple consecutive steps of fine-tuning language models:

\setlist{nolistsep}
\begin{enumerate}[(a)]
\item The decoder is fine-tuned on the training set to predict the change point from a human-written text to a machine-generated text.
\item The decoder makes predictions and outputs the source texts with predicted change points.
\item The first encoder model is fine-tuned on the texts with predicted change points from step (b).
\item The second encoder model is fine-tuned on the mixture of texts from the training set and the texts with predicted change points from step (b).
\item Two encoders are used to predict the indices of change points in test texts.
\item The predicted change points from step (e) are aggregated by averaging.
\end{enumerate}

\begin{figure}[h]
\centering
\includegraphics[width=0.95\linewidth]{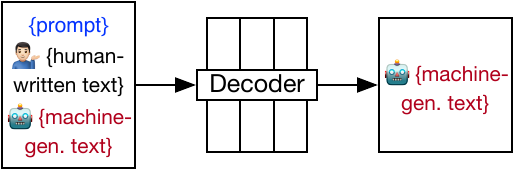}
\caption{We fine-tune the decoder to output only the machine-written text.}
\label{fig:decoder}
\end{figure}

\paragraph{Decoder}\label{par:decoder} The decoder is fine-tuned as follows: the input comprises the prompt and the training text. We experimented with various prompts, including instructing the model to output only the human-written text, the text with an inserted symbol representing the change point, and the machine-generated text alone. Our preliminary experiments suggest that instructing the decoder to output only the machine-generated text yields better results. Therefore, we use this option in subsequent experiments. \autoref{tab:prompt} describes the prompt, and \autoref{fig:decoder} illustrates fine-tuning the decoder. 
The decoder is used in the first step of the AIpom pipeline: we utilize it to generate initial predictions, which are then further processed by two encoders.

After receiving the predicted text from the decoder, we post-process the original training text and insert a special token \texttt{<BREAK>} directly before the first machine-generated word predicted by the decoder. This allows us to pass the prediction further to the encoder.

\begin{table}[]
    \centering
    \begin{tabular}{|p{0.93\columnwidth}|}
    \hline
    \texttt{As an output, write only the machine-generated part of the provided text. Output must start with ``Answer: ''. Separate tokens by `` ''. If the whole text is human-written, output ``None''. Here is the text: {example[``text'']}} \\
    \hline
    \end{tabular}
    \caption{The prompt used for fine-tuning the decoder.}
    \label{tab:prompt}
\end{table}

\begin{figure}[t!]
\centering
\includegraphics[width=0.95\linewidth]{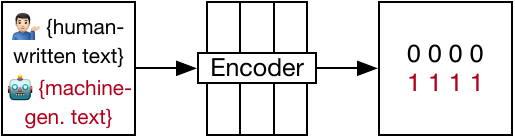}
\caption{We fine-tune the encoder for token labeling. Human-written tokens are assigned zeros, while machine-generated tokens are assigned ones.}
\label{fig:encoder}
\end{figure}

\paragraph{Encoder} The encoder is fine-tuned to label input texts on a token-wise way. Each token in the human-written segment is labeled with a zero, while each machine-generated token is labeled with one. In our final prediction, we determine the position of the word in which the first ``1'' label appears, indicating machine-generated text. See \autoref{fig:encoder} for illustration.

The AIpom pipeline involves fine-tuning two encoders. The first encoder is trained on a dataset consisting of texts labeled by the decoder. In contrast, the second encoder is fine-tuned using a dataset that includes both the decoder's predictions and the original source texts from the training set. we receive the predicted change point from each encoder, which we aggregate by averaging.

\section{Experiments}\label{sec:exps}

\paragraph{Overview} We design a series of experiments using two recent language models, a decoder \textit{Mistral-7B-OpenOrca}\footnote{\href{https://huggingface.co/Open-Orca/Mistral-7B-OpenOrca}{\texttt{hf.co/Open-Orca/Mistral-7B-OpenOrca}}} \cite{jiang2023mistral} and an encoder \textit{DeBERTaV3-Large}\footnote{\href{https://huggingface.co/microsoft/deberta-v3-large}{\texttt{hf.co/microsoft/deberta-v3-large}}} \cite{he2023debertav}, selected based on their performance on standard NLP benchmarks and computational requirements. 

First, we establish a baseline for the decoder model by zero-shot prompting and then compare it to Low-Rank Adaptation (LoRA) tuning \cite{hu2021lora}, which yields significantly better results. Second, we look into improving the performance of the encoder model. We experiment with hyperparameter selection and feeding the encoder model with texts labeled by the decoder model. The combination of the decoder and encoder model outperforms each pipeline component individually.

We only use the development set to evaluate our pipeline and choose our final submission based on the MAE on the development set. In \S\ref{results}, we report the ablation studies results on both development and official test sets\footnote{The shared task organizers have released the gold annotation for the official test set.}.

\paragraph{Decoder fine-tuning and inference}
To fine-tune the Mistral model, we employ LoRA layers tuning with the \texttt{SFTTrainer} class from the transformers library \cite{wolf-etal-2020-transformers}. The model is fine-tuned on to output machine-generated texts, that is the loss functions are computed only on the model-generated parts. We experimented with learning rates in the range $[\texttt{1e-5}, \texttt{5e-5}]$ with an increment of \texttt{1e-5}, and \texttt{warmup\_ratio} $\in \{0.01, 0.03, 0.05\}$. Based on the results observed on the development set, we select a learning rate of \texttt{2e-5}, combined with a \texttt{warmup\_ratio=0.03} and the \texttt{CosineLRScheduler}. For the Parameter-Efficient Fine-Tuning (PEFT) configuration, we adhere to the recommended parameters for Mistral: \texttt{rank=32}, \texttt{lora\_alpha=64}, and \texttt{lora\_dropout=0.05}. The batch size is set to 4 and the model is fine-tuned for 4 epochs.

We fine-tune the model to start its response with the \texttt{"Answer: "} template. This helps improve performance at the inference stage by providing easier-to-clean-up predictions, ensuring they always start the same way. We use the vLLM framework\footnote{\href{https://github.com/vllm-project/vllm}{\texttt{github.com/vllm-project/vllm}}} \cite{kwon2023efficient} for text generation, with default hyperparameters, the sampling temperature of 1, and \texttt{top\_p} of 1.

\paragraph{Data labeling with decoder} To prepare the training set for the encoder, we split the training set into two folds and perform LoRA tuning on two Mistral models with the same hyperparameters on each fold. Then, each fold is labeled using the decoder fine-tuned on the other fold. This helps us track the decoder's performance and reduce overfitting. During testing, we fine-tune another Mistral model on the entire training set and assess its performance on the development set. It is worth noting that we apply the post-processing step described in \S\ref{sec:aipom}, specifically in the \hyperref[par:decoder]{''Decoder``} paragraph, to the predicted text before passing it to the encoder.

\begin{table*}[ht!]
\begin{center}
    \footnotesize
    \centering
\begin{tabular}{@{}llllrr@{}}
\toprule
Setup & Model                  & Fine-tuning setup                     & \texttt{<BREAK>} in the input & Dev MAE         & Test MAE         \\ \midrule
1.    & LoRA Mistral     &   Training set                  &                                                     & 2.41            & 17.00            \\
2.    &  DeBERTa     & Pred. from Mistral             & \multicolumn{1}{c}{\checkmark}       & 1.74            & 17.15            \\
3.    &  DeBERTa     &   Training set + pred. from Mistral & \multicolumn{1}{c}{\checkmark}       & 1.74            & $\mathbf{15.21}$ \\
4.    & AIpom  & 2. + 3.                         & \multicolumn{1}{c}{\checkmark}       & $\mathbf{1.68}$ & 15.94            \\ \midrule
\multicolumn{6}{c}{Ablation experiments} \\
\midrule
5.    & zero-shot Mistral      &   Training set                      &                                                     & 56.51           & 80.81            \\
6.    &  DeBERTa     &   Training set                      &                                                     & 2.15            & 19.97            \\
7.    &  DeBERTa     &   Training set + pred. from Mistral &                                                     & 1.91            & 16.49            \\ \bottomrule
\end{tabular}
\end{center}
\caption{MAE scores on the development and official test sets for different setups and ablation experiments. Setup details include the model used, fine-tuning setup, and presence of \texttt{<BREAK>} in the input data at the inference stage. The top table shows each language model's performance in the AIpom pipeline. The bottom part shows ablation experiments.}
\label{table:main_res}
\end{table*}

\paragraph{Encoder fine-tuning} We build upon the baseline code provided by the task organizers\footnote{\href{https://github.com/mbzuai-nlp/SemEval2024-task8/tree/main/subtaskC/baseline}{\texttt{github.com/mbzuai-nlp/SemEval2024-task8}}}, enhancing it to effectively fine-tuning the DeBERTa model. We explore a range of learning rates $[\texttt{1e-5}, \texttt{5e-5}]$ with an increment of \texttt{1e-5} to identify the optimal value for fine-tuning our model. Our final fine-tuning strategy utilizes the Adam optimizer \cite{ADAM}, with a learning rate of \texttt{3e-5} and the default \texttt{CosineLRScheduler}. To ensure consistency across all experiments, we use a maximum sequence length of 1024 for text tokenization, maintain a constant batch size of 64, and limit the maximum number of epochs to 6. To reduce overfitting, we freeze a certain number of bottom DeBERTa layers. Specifically, we experiment with fine-tuning only the top $N \in \{6, 12, 18\}$ layers out of the total 24. Through experiments, we determine that fine-tuning only the top 12 layers produces the best results.

\paragraph{Hardware specification}  We run experiments on a single GPU TESLA A100 80 GB. Model fine-tuning is conducted using the \texttt{transformers} library. The fine-tuning for DeBERTa requires approximately 3.5 hours to complete, while the inference on the official test set runs within 15 minutes. The LoRA tuning for Mistral lasts approximately 12 hours, with the inference on the official test set taking a few hours. To speed up the prediction phase, we employ the vLLM framework, designed specifically for optimizing the inference. This implementation significantly reduces inference time, with the official test set predictions generated in just 30 minutes.

\section{Results}

\label{results}
\autoref{table:main_res} provides a detailed comparison of the performance metrics across different models and experimental setups. The key results are:

\begin{itemize}
\item Fine-tuning Mistral with LoRA tuning (setup 1) on the training set outperforms zero-shot prompting (setup 5) by a wide margin.
\item Fine-tuning DeBERTa using the Mistral predictions (setup 2) leads to higher results than fine-tuning DeBERTa on the training set (setup 6).
\item Adding a \texttt{<BREAK>} token to the input at the inference stage improves the performance of the DeBERTa model (setup 3 vs. setup 7).
\item Averaging predictions of two DeBERTa models (setup 4) leads to the best results on the development set.
\end{itemize}

The overall best results on the official test set are achieved with setup 3, where the DeBERTa model is fine-tuned on a dataset consisting of both the training set and predictions from the Mistral model, and the \texttt{<BREAK>} token is added to the input at the inference stage.\footnote{These results are achieved after the shared task submission deadline and hence not submitted for the official evaluation stage.}

\paragraph{Decoder vs. encoder}  Fine-tuned encoder models exhibit inferior performance compared to LoRA-tuned decoder. Specifically, the decoder struggles to comprehend the task in a zero-shot setting, evidenced by a high MAE of 80.81 on the official test set. However, with LoRA tuning, the decoder achieves a significantly lower MAE of 17.00, outperforming the single encoder model's MAE of 19.97. The encoder models are adaptable and integrate diverse inputs, including prompts and predictions from additional decoder models. 

\paragraph{Benefits of pipelining} We hypothesize that encoder models benefit from integrating inputs from a decoder. Our pipeline yields the final MAE of 15.94 while fine-tuning only the single encoder model results in a higher MAE score of 19.97.

\paragraph{Robustness}

While averaging predictions helps improve the overall performance, we find that the AIPom's robustness has room for improvement. In particular, we observe the performance decrease when comparing the results on the development and official test sets. Setup 3, with an MAE of 1.74 on the development set, performs better than an MAE of 15.21 on the official test set. At the same time, Setup 4 (our final submission) achieves a slightly better development MAE but a worse MAE on the official test set. Setup 3 involves finetuning on a mixture of data, showing how using more data can boost the performance and improve the robustness, especially when the dataset is small. We leave improving the out-of-domain robustness for future work.

\section{Conclusion}
This paper presents the AIpom system submitted to  SemEval-2024 Task 8. Our solution achieves 2nd place out of 33 participating teams in Subtask C. We introduce a novel method that utilizes a pipeline of decoder and encoder models. The advantage of this approach is that the models are exposed to both the original data and the predictions from previous steps. We believe this approach holds significant potential, as it allows for creating pipelines comprising various models, mimicking the transfer of learned knowledge. We plan to further improve our system by exploring different combinations of models and longer pipelines. Additionally, we aim to enhance the system's robustness to handle domain-shift scenarios.

\section*{Acknowledgements}
AS's work results from a research project implemented in the Basic Research Program at the National Research University Higher School of Economics (HSE University). 
We acknowledge the computational resources of HSE University's HPC facilities.

\bibliography{custom,anthology}

\end{document}